\documentclass[english,runningheads,a4paper]{llncs}
\usepackage{graphicx}
\usepackage{url}
\usepackage{multirow}
\usepackage{float}
\usepackage{tabularx}
\usepackage{makecell}
\usepackage{tabulary}
\usepackage{hyperref}
\usepackage{hhline}
\usepackage{array}
\usepackage{amsmath}
\usepackage{fancyhdr}
\usepackage{soul}
\usepackage{color}

\usepackage{subcaption}
\usepackage{pgffor}
\usepackage{xparse}
\usepackage{ifthen}

\usepackage[symbol]{footmisc}

\newcolumntype{P}[1]{>{\centering\arraybackslash}p{#1}}

\DeclareDocumentEnvironment{env_comparison}{O {10} m}
{   
    \begin{figure}[!htb]
    \centering
    \begin{subfigure}[b]{.32\linewidth}
    \includegraphics[width=\linewidth]{figures/suppl/base_full_res/depth_full/#1.png}
    \caption{RTPose\_640x480}
    \end{subfigure}
    \begin{subfigure}[b]{.32\linewidth}
    \includegraphics[width=\linewidth]{figures/suppl/base_80x60/depth_80x60/#1.png}
    \caption{RTPose\_80x60}
    \end{subfigure}
    \begin{subfigure}[b]{.32\linewidth}
    \includegraphics[width=\linewidth]{figures/suppl/base_64x48/depth_64x48/#1.png}
    \caption{RTPose\_64x48}
    \end{subfigure}\\
    \begin{subfigure}[b]{.32\linewidth}
    \includegraphics[width=\linewidth]{figures/suppl/gt/color/#1.png}
    \caption{GT}
    \end{subfigure}
    \begin{subfigure}[b]{.32\linewidth}
    \includegraphics[width=\linewidth]{figures/suppl/sr02_80x60/depth_80x60/#1.png}    
    \caption{DepthPose\_80x60}
    \end{subfigure}    
    \begin{subfigure}[b]{.32\linewidth}
    \includegraphics[width=\linewidth]{figures/suppl/sr02_64x48/depth_64x48/#1.png}    
    \caption{DepthPose\_64x48}
    \end{subfigure}
    \caption{}
    \label{fig:comp64x48}
    \end{figure}
}
{}

\newcommand{\compare}[1]
{
    \begin{env_comparison}[#1]{}
    \end{env_comparison}
}

\begin{document}
\title{Human Pose Estimation on Privacy-Preserving Low-Resolution Depth Images}
\titlerunning{Human Pose Estimation on Low-Resolution Depth Images}

\author{Vinkle Srivastav\inst{1} \and Afshin Gangi\inst{1,2} \and Nicolas Padoy\inst{1}}
\authorrunning{Srivastav et al.}

%

\urldef{\mailsa}\path{{srivastav | padoy}@unistra.fr}
\institute{ICube, University of Strasbourg, CNRS, IHU Strasbourg, France\\ \mailsa
  \and Radiology Department, University Hospital of Strasbourg, France}
 
%
\maketitle              
\begin{abstract}
Human pose estimation (HPE) is a key building block for developing AI-based context-aware systems inside the operating room (OR). The 24/7 use of images coming from cameras mounted on the OR ceiling can however raise concerns for privacy, even in the case of depth images captured by RGB-D sensors. Being able to solely use low-resolution privacy-preserving images would address these concerns and help scale up the computer-assisted approaches that rely on such data to a larger number of ORs. In this paper, we introduce the problem of HPE on low-resolution depth images and propose an end-to-end solution that integrates a multi-scale super-resolution network with a 2D human pose estimation network. By exploiting intermediate feature-maps generated at different super-resolution, our approach achieves body pose results on low-resolution images (of size 64x48) that are on par with those of an approach trained and tested on full resolution images (of size 640x480).
\end{abstract}
\footnotetext{Published at International Conference on Medical Image Computing and Computer-Assisted Intervention - MICCAI 2019.}

\section{Introduction}

\begin{figure}[htb]
	\centering
	\begin{subfigure}[t]{1.5in}
		\includegraphics[width=1.5in]{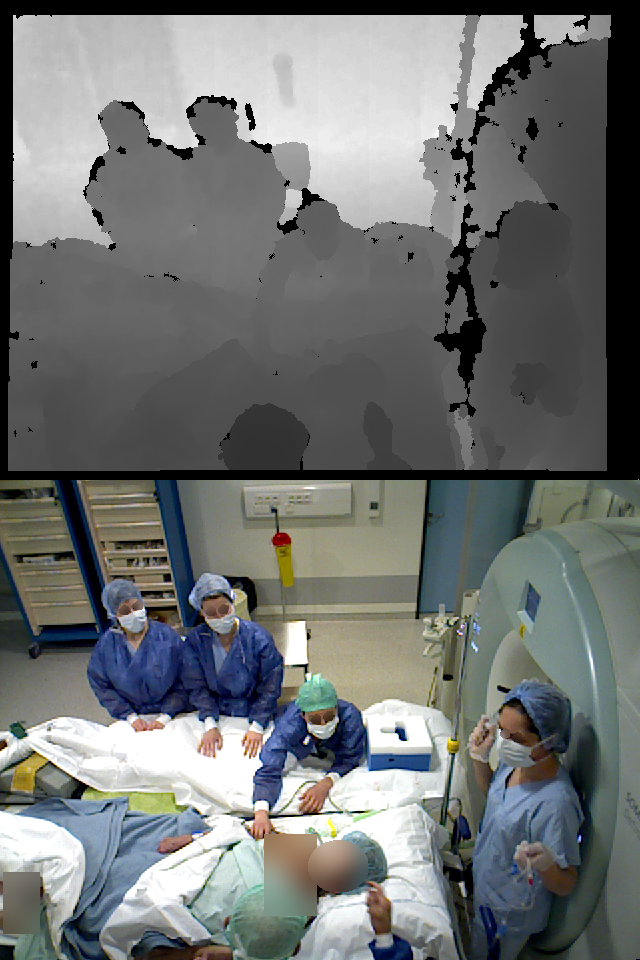}
		\caption{640x480 (1x)}
	\end{subfigure}	
	\begin{subfigure}[t]{1.5in}
		\includegraphics[width=1.5in]{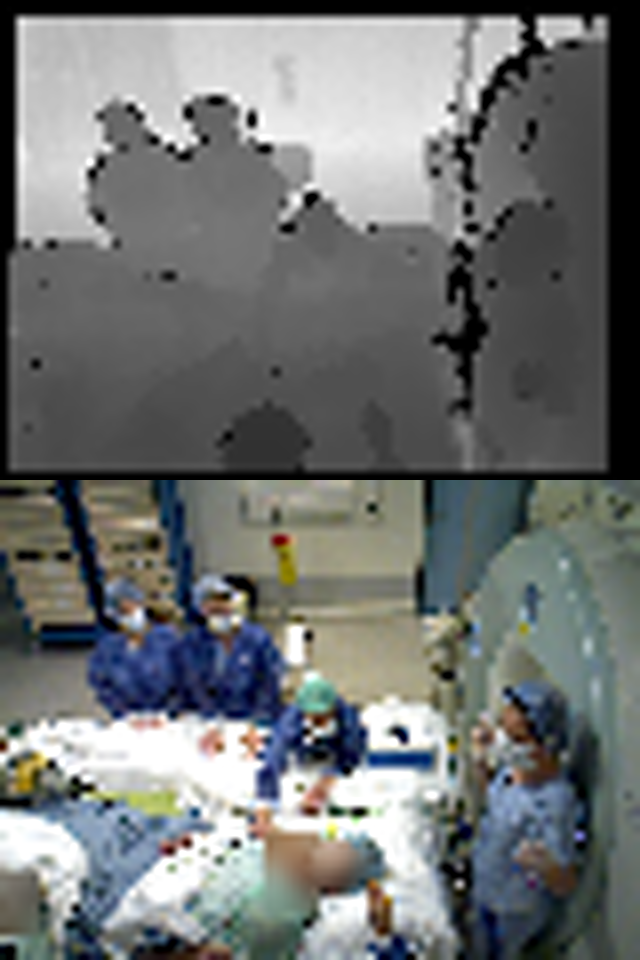}
		\caption{80x60 (8x)}
	\end{subfigure}	
	\begin{subfigure}[t]{1.5in}
		\includegraphics[width=1.5in]{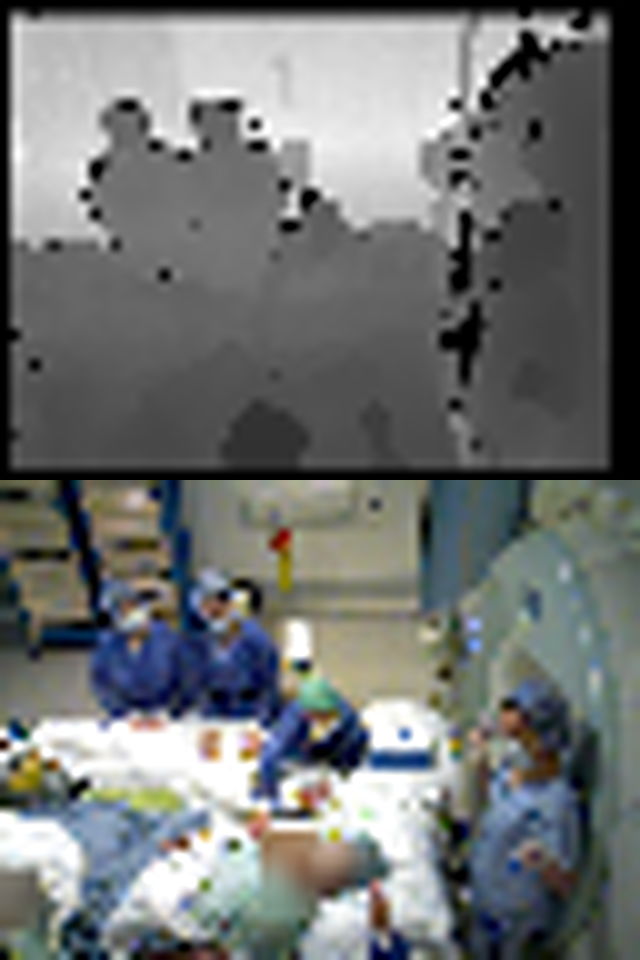}
		\caption{64x48 (10x)}
	\end{subfigure}			
	\caption{Depth and color images from MVOR dataset \cite{srivastav2018mvor} down-sampled at different resolutions using bicubic interpolation (resized for better visualization). Low-resolution depth images contain little information for the identification of patient and health professionals. Corresponding color images in the second row are shown for better appreciation of the downsampling process.}
	\label{depth_imgs}
\end{figure}

\noindent Modern hospitals could highly benefit from the use of smart assistance systems that are able to support the workflow by exploiting digital data from equipment and sensors through artificial intelligence and surgical data science \cite{maier2017surgical,padoy2019machine}. This is illustrated by the recent development of new applications, such as patient activity monitoring inside intensive care units (ICU) \cite{ma2017measuring}, 
staff hand-hygiene recognition \cite{haque2017}, radiation exposure monitoring during hybrid surgery \cite{rodas2017see} and workflow steps recognition in the operating room (OR) \cite{twinanda2016multi}.

These systems, which have a huge potential to improve safety and care, all rely on machine intelligence using computer vision models to extract semantic information from visual data. In particular, human detection and pose estimation in the operating room \cite{belagiannis2016parsing,haque2016towards,kadkhodamohammadi2017-ar} is one of the key components to develop such applications. 
Constant monitoring by the use of cameras raises however potential concerns for the privacy of patients and health professionals. 
Cameras usually capture the color, depth or both types of images for visual processing. Color images appear to be the most privacy-intrusive, but even textureless depth images can also intrude the privacy when used at sufficiently high-resolution \cite{cheng20173d,haque2018activity}. This is particularly relevant in environments where the number of persons is limited and where the persons could potentially be more easily identified. Figure \ref{depth_imgs} shows depth images at different resolutions. It suggests that low-resolution images could be used for more privacy-compliant computer-vision applications and that their recording  could be better accepted by  clinical institutions. In \cite{haque2018activity}, it has been shown that activity recognition can be performed on low-resolution depth images captured for the tasks of hand-hygiene classification and ICU activity logging. In this work, we investigate whether low-resolution depth images contain sufficient information for accurate human pose estimation (HPE).

HPE consists of localizing human keypoints in images. Methods for human pose estimation are different for color and depth images both in terms of model architectures and complexity of training datasets. In the case of color images, deep learning models have recently shown remarkable progress with the help of large scale \textit{in the wild} annotated datasets, such as COCO \cite{lin2014microsoft}.
Deep learning models for HPE can generally be grouped into bottom-up and top-down approaches. Bottom-up approaches first detect the keypoints and then group them to form skeletons \cite{cao2016realtime}, whereas top-down approaches first detect the person using person detectors and then use single person pose estimator to estimate body joints in each detected box \cite{xiao2018simple}. Top-down approaches are often more accurate due to their two-stage design but slower in comparison to bottom-up approaches. For depth images, Shotton et al. \cite{shotton2011real} use high-resolution synthetic depth dataset to train the models, while Haque et al. \cite{haque2016towards} focus on single person pose estimation using datasets recording actors performing simulated actions. Recently, Srivastav et al. \cite{srivastav2018mvor} have introduced the MVOR dataset, which contains color and depth images captured in the OR along with ground truth human poses. They have also evaluated recent HPE methods. This is therefore an interesting testbed for multi-person pose estimation on depth data captured during real surgical activities, which we will use in this work.

Current methods for HPE inside the OR have been developed using standard resolution images \cite{belagiannis2016parsing,kadkhodamohammadi2017-ar}. We have found that state-of-the-art models, which are trained on the high-resolution images, perform poorly on the corresponding 
low-resolution images. In this paper, we therefore propose an approach for the human pose estimation problem on low-resolution depth images. To the best of our knowledge, this is the first work that attempts to solve this task. 

To train our system, we use a non-annotated dataset of synchronized RGB-D images captured in the OR environment. Unlike conventional approaches, which use either manual or synthetically rendered annotations challenging to generate, we propose to use the detections from a state-of-the-art method applied to the color images as pseudo ground truth for the corresponding depth images. This simple idea turns out to be very effective. Indeed, as our approach only requires a set of RGB-D images at train time, it can be easily retrained in any facility since no annotation process is needed. Then, it can run round the clock on low-resolution depth images from the same facility.
Our HPE approach is a network which integrates super-resolution modules with a 2D multi-person body keypoint estimator based on RTPose \cite{cao2016realtime}. It utilizes intermediate super-resolution feature maps to better learn the high-frequency features. With the proposed architecture, we achieve the same results as a network trained on the standard resolution images and improve by $6.5$\% the results of a baseline method which up-samples the low-resolution images with bicubic interpolation before feeding them to the pose estimation network.


\section{Methodology}

\begin{figure}[tb]
\includegraphics[width=1.0\linewidth]{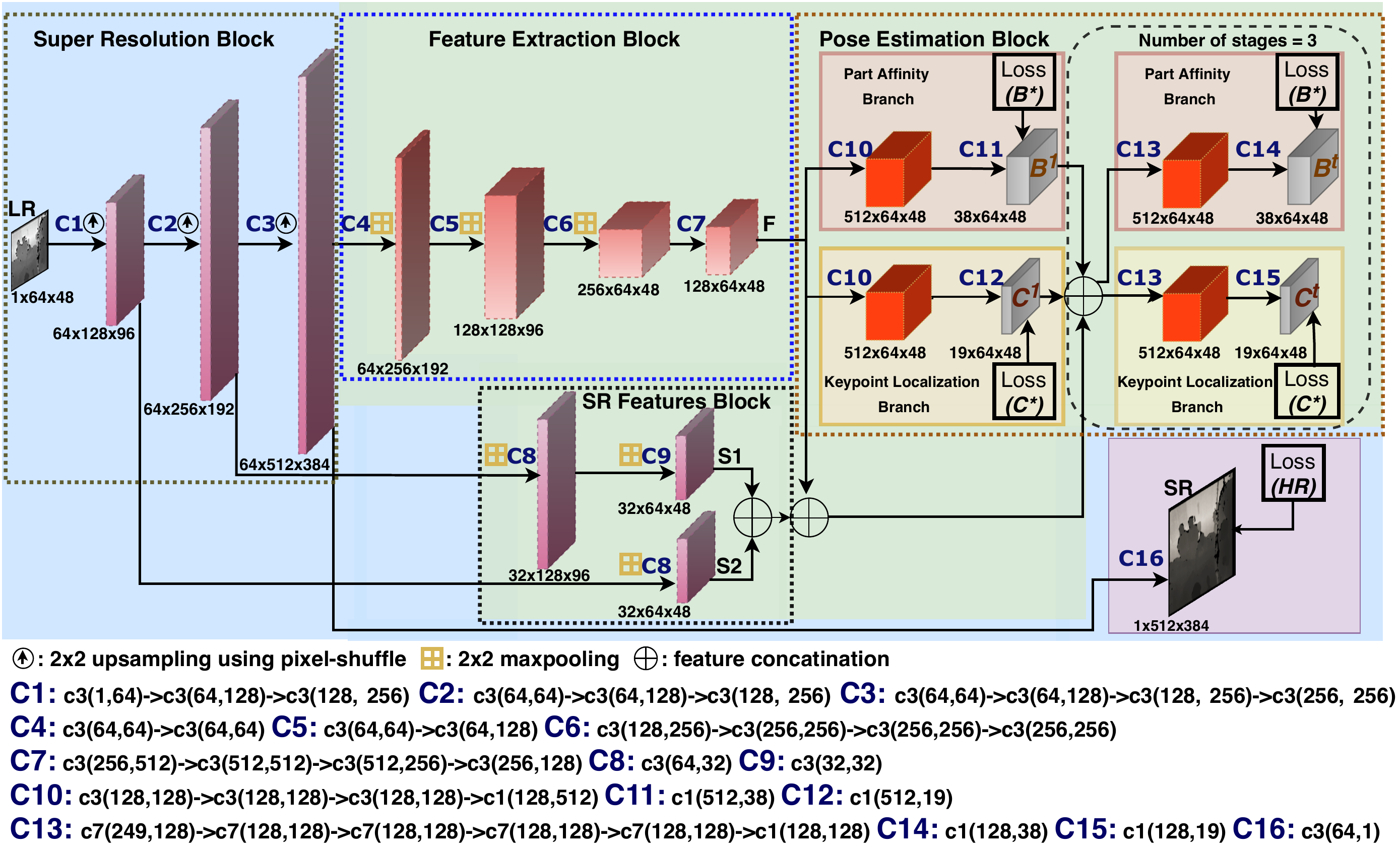} 
\caption{Proposed architecture. The super-resolution block increases the spatial resolution by a factor of 8x and generates intermediate SR feature maps (S1, S2) used by the pose estimation block to learn high-frequency features. All losses are mean square error losses. C1 to C16 are convolution layers grouped together for better visualization and described below the figure, where c1(n1,n2), c3(n1,n2), c7(n1,n2) each represent a convolution layer with kernel size 1x1, 3x3, 7x7 and padding 0, 1, 3, respectively. Parameters n1 and n2 are the numbers of input and output channels and all convolution layers are followed by RELU non-linearity.}
\label{fig:architecture}
\end{figure}

\subsection{Architecture}
Our approach is inspired by the recent developments in the area of super-resolution and multi-person human pose estimation. We propose to integrate a super-resolution image estimator and a 2D multi-person pose estimator in a joint architecture, illustrated in Figure \ref{fig:architecture}.
This architecture is based on modification from the RTPose network \cite{cao2016realtime}. Besides yielding competitive results on COCO and MVOR, RTPose has the advantage to perform multi-person pose estimation in a single step, thereby simplifying the integration and training of the super-resolution modules. It is composed of a \texttt{feature extraction block} and a \texttt{pose estimation block} shown in Figure \ref{fig:architecture}.
We introduce a \texttt{super-resolution block}, which does not only increase the spatial resolution but also generates super-resolution (SR) feature maps (S1, S2). These intermediate feature-maps contain high-frequency details, which are lost during the low-resolution (LR) image generation process and used in the \texttt{pose estimation block} for better localization. 
The \texttt{super-resolution block} uses a multi-stage design, where each stage increases the spatial resolution of the features maps by a factor of two using the pixel-shuffle algorithm \cite{shi2016real} (while reducing the number of channels by four). During training, a complete SR image is generated to compute the auxiliary loss \textbf{L\_HR}, which compares the SR image to the ground truth high-resolution (HR) depth image using the L2 norm. This helps to train the \texttt{super-resolution block} and refines the input to the \texttt{SR features block}. Note that during training, errors from the pose estimation are also back-propagated to these blocks. Furthermore, at test time only LR images are used and no SR images need to be generated by the network since only the SR feature maps are used. 

RTPose was originally developed for color images. Since depth images contain fewer texture details, we have made the architecture more computationally efficient by reducing the number of iterative refinement stages from five to three. The network uses two separate branches, one for keypoint localization and another to compute part affinity maps \cite{cao2016realtime}. In our architecture, these two branches consume the 3 types of features (F, S1, S2), where F are the features extracted from the high-resolution feature maps provided by the super-resolution block. The final skeleton is generated from the part affinity and keypoint localization heatmaps using the bipartite graph matching algorithm presented in \cite{cao2016realtime}. 
Losses in the pose estimation network are used as in \cite{cao2016realtime}, but now take the input from the SR feature maps (S1, S2). At each stage $t$, two L2 losses $L\_B^t$ and $L\_C^t$ are computed from the predicted part affinity/keypoint localization heatmaps ($B^t$/$C^t$) and the ground truth heatmaps ($B^*$/$C^*$). All the $L\_B^t$ and $L\_C^t$ losses are summed together to form the pose estimation loss \textbf{L\_P}.
Finally, the total loss is the sum of \textbf{L\_HR} and \textbf{L\_P}. We have chosen to weigh both terms equally as we observe that their magnitudes are similar. The complete network is trained end-to-end jointly for both super-resolution and pose estimation.

\subsection{Ground-truth generation}
\label{sec:gt}
In the literature, authors have either used manually annotated or synthetically generated datasets to train for HPE on depth images. Manual annotations can be expensive and time-consuming, and synthetic annotations are difficult to generate due to the constraint of realistic rendering and do not always generalize well to real scenarios. Therefore, we use an alternate approach to generate annotations. This approach is based on the observation that the RGBD cameras capture synchronized color and depth streams, and recent HPE methods trained on the COCO dataset \cite{lin2014microsoft} work remarkably well on the color images. Therefore, we use detections from the color images to train the model for the depth images. To facilitate this approach, we collected an unlabeled RGBD dataset containing synchronized 80k color and depth images captured in the OR during real surgical procedures. Then, we used the state-of-art person detector Mask-RCNN \cite{he2017mask} and a single person pose estimator MSRA \cite{xiao2018simple} on color images to generate detections. 
We filter out the false positives and retain high-quality detections in both the stages using thresholds selected from the qualitative results on a small set of images. 
This approach generates pseudo ground truth automatically without using any human annotation efforts. It is therefore scalable and can be deployed to any facility. For human pose estimation, we choose here a two steps method based on Mask-RCNN and MSRA for their state-of-the-art performance on the public COCO dataset. Note that such a two-step method would be less convenient to use in our approach, due to the large architectures involved and the fact that super-resolution would need to be integrated into both. 



\section{Experiments and Results}

\subsubsection{Training setup:} 
We use the dataset of 80k images and the pseudo ground truth described in Section \ref{sec:gt} for training.  It contains 20k images from four categories, where each category includes images with one, two, three and four or more persons. We split the dataset into 77k training and 3k validation images. When downsampling the images to sizes 80x60 and 64x48, we use bicubic interpolation. To generate pseudo ground truth, we use a threshold of 0.7 in the person-detector stage and then select the skeleton if at least 4 keypoints are detected with a score greater than 0.35.
We use PyTorch deep learning framework in our experiments. The depth images are normalized in the range [0, 255] and we train our networks using the stochastic gradient descent optimizer with a momentum of 0.9. The initial learning rate is set to 0.001 with a step decay of 0.1 after 12k iterations and each model is trained for 32k iterations with a batch size of 12. We use the pre-trained weights from the authors of RTPose to initialize the pose-estimator networks. Note that these weights were originally obtained using the color images from the COCO dataset. For the layers that have been modified in the pose-estimation network and contain a larger number of channels (e.g. to accommodate S1 and S2), we repeated the same weights and perturbed them by a small random number. The weights of the super-resolution network are initialized using orthogonal initialization \cite{saxe2013exact}.


\subsubsection{Testing setup:} 
We evaluate our method on the publicly available depth images of the MVOR dataset \cite{srivastav2018mvor}, which contains images of size 640x480 captured in an OR from 3 different viewpoints during actual clinical interventions. The training dataset comes from the same environment and camera setup but contains data captured on different days. During testing, we use the flip-test, namely average the original heatmaps with the heatmaps obtained after flipping the images horizontally to refine the predictions. We use the percentage of correct keypoints (PCK) \cite{yang2013articulated} as an evaluation metric, which is widely used to measure the localization accuracy of the detected skeletons in multi-person scenarios.

\subsection {Results}
We show our results in Table \ref{table:results_main}. RTPose\_640x480, RTPose\_80x60, and RTPose\_64x48 are baseline RTPose models that do not use any super-resolution and are trained on 640x480 (full-size), 80x60, and 64x48 size depth images, respectively. These RTPose variants are the original models modified to take a 1-channel input. 
The degraded 80x60 and 64x48 images are resampled to the original size using bicubic interpolation to match the input size of the network. 
DepthPose\_80x60 and DepthPose\_64x48 are our proposed networks directly trained on 80x60 and 64x48 low-resolution images.
Results show that the DepthPose\_64x48 network, which uses 10x downsampled images, performs on par with the baseline trained on full-size image. Accuracy is improved by over 6.5\% compared to the baseline RTPose\_64x48. DepthPose\_80x60 performs even better than RTPose\_640x480 (an interesting fact also observed in \cite{haque2018activity} in the context of activity classification) and is 3.6\% better than RTPose\_80x60. 

\begin{figure}[htb]
	\centering
	\begin{subfigure}[t]{2.0in}
		\includegraphics[width=2.0in]{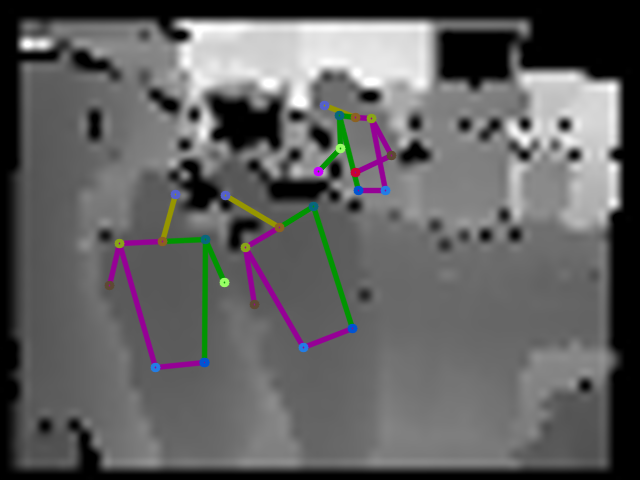}
	\end{subfigure}	
	\begin{subfigure}[t]{2.0in}
		\includegraphics[width=2.0in]{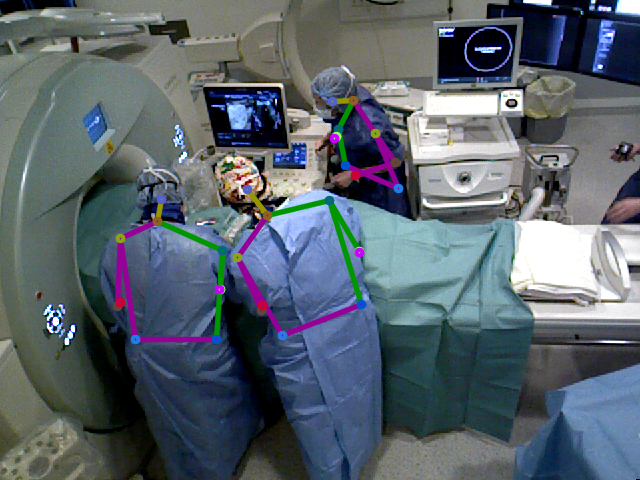}
	\end{subfigure}	
	\caption{Qualitative results of the DepthPose\_64x48 model on a 64x48 LR depth image with 3 persons. Ground truth is overlaid on the color images for better visualization.}
    \label{fig:qualitative_results}
	\label{depth_imgs}
\end{figure}

We have also evaluated the quality of the pseudo ground truth by running the Mask-RCNN and MSRA models on the color images from MVOR. The resulting PCK value is 76.2, showing that there still exists a gap of around 9\% to be filled between the depth and color images. This may also explain the improved results of DepthPose\_80x60 model, which takes advantage of an improved architecture compared to the full-size RTPose\_640x480 model. 
Figure \ref{fig:qualitative_results} shows some qualitative results of the DepthPose\_64x48 model. Additional qualitative comparisons are available in the supplementary material. 

\begin{table}[tb]
    \begin{center}
    \begin{tabular}{|P{2.8cm}|P{1.3cm}|P{1.3cm}|P{1.3cm}|P{1.3cm}|P{1.3cm}||P{1.3cm}|}
    \hline
     & Head & Shoulder & Hip & Elbow & Wrist & Average \\
    \hline
    RTPose\_640x480 & 82.9 & 82.2  & 57.0 & 68.5  & 42.8  & 66.7 \\
    \Xhline{4\arrayrulewidth}
    RTPose\_80x60 & 81.1 & 80.0 & 54.7 & 65.3  & 37.3 & 63.7  \\
    \hline
    RTPose\_64x48 & 77.8 & 76.4 & 52.9 & 60.7 & 32.0 & 60.0  \\
    \Xhline{4\arrayrulewidth}
    DepthPose\_80x60 & 84.3 & 83.8 & 55.3 & 69.9 & 43.3 & 67.3  \\
    \hline
    DepthPose\_64x48 & 84.1 & 83.4 & 54.3 & 69.0 & 41.4 & 66.5  \\
    \Xhline{4\arrayrulewidth}
    SR+RTPose\_80x60 & 83.5 & 82.7 & 54.1 & 68.1 & 40.5 & 65.8  \\
    \hline
    SR+RTPose\_64x48 & 82.5 & 81.3 & 51.0 & 66.3 & 37.8 & 63.8  \\
    \hline
    \end{tabular}
    \end{center}
    \caption{Results of our proposed method (DepthPose) compared to the baselines (RTPose and SR+RTPose) for different image resolutions.}
    \label{table:results_main}
\end{table}
\subsubsection {Comparative study without SR feature maps:}
We also experiment to better understand the effect of using super-resolution. Instead of giving to the baselines RTPose\_80x60 and RTPose\_64x48 images that are up-sampled with bicubic interpolation, we feed and train these networks with images up-sampled separately using a super-resolution network. The super-resolution network corresponds to the super-resolution block trained independently using loss L\_HR.  We observe in Table \ref{table:results_main}  that this procedure (SR+RTPose) improves the overall accuracy, but yields result inferior to DepthPose by 1.5\% and 2.7\% for 80x60 and 64x48 images, respectively. This shows that the use of intermediate SR feature maps in the pose estimation network helps to better localize keypoints. Also, SR+RTPose has the disadvantage to explicitly generate super-resolution images, the privacy compliance of which would need to be considered.

\section{Conclusion} 
In this paper, we present an approach for high-resolution multi-person 2D pose estimation from low-resolution depth images. Our evaluation on the public MVOR dataset shows that even with a 10x subsampling of the depth images, our method achieves results equivalent to a pose estimator trained and tested on the original-size images. Furthermore, we show that by exploiting high-quality pose detections on the color images of a non-annotated RGB-D dataset, we can generate pseudo ground truth for the depth images and train a decent OR pose estimator. These results suggest the high potential of low-resolution images for scaling up and deploying privacy-preserving AI assistance in hospital environments.

\subsubsection{Acknowledgements}
This work was supported by French state funds managed by the ANR within the Investissements d'Avenir program under references ANR-16-CE33-0009 (DeepSurg), ANR-11-LABX-0004 (Labex CAMI) and ANR-10-IDEX-0002-02 (IdEx Unistra). The authors would also like to thank the members of the Interventional Radiology Department at University Hospital of Strasbourg for their help in generating the dataset.


\clearpage
\section*{\centering *** Supplementary Material ***}
\subtitle{ *** Supplementary Material ***}

\thispagestyle{empty}
The pictures below show additional qualitative results of our proposed models, namely DepthPose\_80x60 and DepthPose\_64x48, w.r.t the baseline models RTPose\_640x480, RTPose\_80x60, and RTPose\_64x48. We also show the ground truth (GT) on color images for better appreciation of the qualitative results. These results show that DepthPose\_80x60 and DepthPose\_64x48 perform better for removing false positives and spurious detections and improve the part localization (see red and green arrows in the figures).\\
\foreach \n in {71,72,176,161,151,143,131,48,18}{\compare{\n}}

\end{document}